\documentclass[11pt,a4paper]{article}

\usepackage{PRIMEarxiv}

\usepackage[utf8]{inputenc} 
\usepackage[T1]{fontenc}    
\usepackage{hyperref}       
\usepackage{url}            
\usepackage{booktabs}       
\usepackage{amsfonts}       
\usepackage{nicefrac}       
\usepackage{microtype}      
\usepackage{fancyhdr}       
\usepackage{graphicx}       
\usepackage{mathtools, amsmath, amsthm, amsfonts, amssymb} 
\usepackage{bm}
\usepackage{cleveref}
\crefname{section}{§\hspace{-0.1cm}}{§§}
\Crefname{section}{§}{§§}
\usepackage{tikz} 
\usetikzlibrary{arrows.meta} 
\usetikzlibrary{positioning, shapes.multipart, arrows}
\usepackage{comment} 

\newenvironment{myquote}[1]%
  {\list{}{\leftmargin=#1\rightmargin=#1}\item[]}%
  {\endlist}

\usepackage{xspace}
\newcommand\Facts{\texttt{Scientific Facts}}
\newcommand\NegFacts{\texttt{NegFacts}\xspace}
\newcommand\Animals{\texttt{Animals}\xspace}
\newcommand\Capitals{\texttt{Capitals}\xspace}
\newcommand\Cities{\texttt{Cities}\xspace}
\newcommand\Elements{\texttt{Elements}\xspace}
\newcommand\Inventions{\texttt{Inventions}\xspace}
\newcommand\Companies{\texttt{Companies}\xspace}
\newcommand\NegCompanies{\texttt{NegCompanies}\xspace}

\usepackage{todonotes}
\usepackage{natbib}
\setcitestyle{aysep={}}
\usepackage[toc,page]{appendix} 
\title{Still No Lie Detector for Language Models: Probing Empirical and Conceptual Roadblocks
}

\author{
  B.A. Levinstein \\
  University of Illinois at Urbana-Champaign \\
  \texttt{benlevin@illinois.edu} \\
   \And
  Daniel A. Herrmann \\
  University of California, Irvine \\
  \texttt{daherrma@uci.edu} \\
}

\begin{document}
\maketitle

\begin{abstract}
We consider the questions of whether or not large language models (LLMs) have beliefs, and, if they do, how we might measure them. First, we evaluate two existing approaches, one due to \cite{azaria2023internal} and the other to \cite{burns2022discovering}. We provide empirical results that show that these methods fail to generalize in very basic ways. We then argue that, even if LLMs have beliefs, these methods are unlikely to be successful for conceptual reasons. Thus, there is still no lie-detector for LLMs. After describing our empirical results we take a step back and consider whether or not we should expect LLMs to have something like beliefs in the first place. We consider some recent arguments aiming to show that LLMs cannot have beliefs. We show that these arguments are misguided. We provide a more productive framing of questions surrounding the status of beliefs in LLMs, and highlight the empirical nature of the problem. We conclude by suggesting some concrete paths for future work.

\end{abstract}

\begin{myquote}{0.1in}
    \textit{One child says to the other “Wow! After reading some text, the AI understands what water is!”
    \ldots \\
    The second child says “All it understands is relationships between words. None of the words connect to reality. It doesn’t have any internal concept of what water looks like or how it feels to be wet. \ldots”\\
    \ldots \\
    Two angels are watching [some] chemists argue with each other. The first angel says “Wow! After seeing the relationship between the sensory and atomic-scale worlds, these chemists have realized that there are levels of understanding humans are incapable of accessing.”
    The second angel says “They haven’t truly realized it. They’re just abstracting over levels of relationship between the physical world and their internal thought-forms in a mechanical way. They have no concept of [*****] or [*****]. You can’t even express it in their language!”} \\
    --- Scott Alexander, \textit{Meaningful}
\end{myquote}

\newpage

\section{Introduction}

Do large language models (LLMs) have beliefs? And, if they do, how might we measure them? 

These questions have a striking resemblance to both philosophical questions about the nature of belief in the case of humans (\cite{ramsey2016truth}) and economic questions about how to measure beliefs (\cite{savage1972foundations}).\footnote{\cite{diaconis2018ten} give a concise and thoughtful introduction to both of these topics.} 

These questions are not just of intellectual importance but also of great practical significance. It is news to no one that LLMs are having a large effect on society, and that they will continue to do so. Given their prevalence, it is important to address their limitations. One important problem that plagues current LLMs is their tendency to generate falsehoods with great conviction. This is sometimes called \textit{lying} and sometimes called \textit{hallucinating} \citep{ji2023survey, evans2021truthful}. One strategy for addressing this problem is to find a way to read the beliefs of an LLM directly off its internal state.  Such a strategy falls under the broad umbrella of model interpretability,\footnote{See \cite{lipton2018mythos} for a conceptual discussion of model interpretability.} but we can think of it as a form of mind-reading with the goal of detecting lies. 

Detecting lies in LLMs has many obvious applications. It would help us successfully deploy LLMs at all scales: from a university student using an LLM to help learn a new subject, to companies and governments using LLMs to collect and summarize information used in decision-making. It also has clear applications in various AI safety research programs, such as Eliciting Latent Knowledge (\cite{ELK}).

In this article we tackle the question about the status of beliefs in LLMs head-on. We proceed in two stages. First, we assume that LLMs \textit{do} have beliefs, and consider two current approaches for how we might measure them, due to \cite{azaria2023internal} and \cite{burns2022discovering}. We provide empirical results that show that these methods fail to generalize in very basic ways. We then argue that, even if LLMs have beliefs, these methods are unlikely to be successful for conceptual reasons. Thus, \textit{there is still no lie-detector for LLMs}. 

After describing our empirical results we take a step back and consider whether or not we should expect LLMs to have something like beliefs in the first place. We consider some recent arguments aiming to show that LLMs cannot have beliefs (\cite{bender2021dangers}; \cite{shanahan2022talking}). We show that these arguments are misguided and rely on a philosophical mistake. We provide a more productive framing of questions surrounding the status of beliefs in LLMs. Our analysis reveals both that there are many contexts in which we should expect systems to track the truth in order to accomplish other goals but that the question of whether or not LLMs have beliefs is largely  an empirical matter.\footnote{We provide code at \url{https://github.com/balevinstein/Probes}.}

\section{Overview of Transformer Architecture}
\label{overview}

\begin{figure}
\centering
\begin{tikzpicture}[
  every node/.style={font=\rmfamily, align=center, inner sep=0.2em, scale=0.8},
  layer/.style={draw, rounded corners, minimum width=1.3cm, minimum height=1.3cm},
  arrow/.style={->, shorten >=1pt, >=stealth, line width=0.7pt},
  token/.style={circle, draw, minimum size=0.5cm}
]

\foreach \word [count=\i] in {\texttt{Mike},\texttt{Trout},\texttt{plays},\texttt{for},\texttt{the}} {
  \node (input-\i) at (\i * 1.5, 0) {\word};
}

\foreach \i in {1,...,5} {
  \foreach \j [count=\k] in {1,...,5} {
    \node[layer, fill=orange!20] (hidden-\i-\k) at (\k * 1.5, \i * 1.4) {$\langle \textcolor{gray}{\tiny\bullet}, \textcolor{gray}{\tiny\bullet}, \textcolor{gray}{\tiny\bullet} \rangle$};
  }
}

\foreach \i in {1,...,5} {
  \node[above=0.3cm of hidden-5-\i] (output-\i) {$p_\i$};
}

\foreach \i in {1,...,5} {
  \foreach \j in {1,...,5} {
    \ifnum\i=1
      \draw[arrow] (input-\j) -- (hidden-\i-\j);
    \fi

    \ifnum\i<5
      \pgfmathtruncatemacro\nexti{\i+1}
      \draw[arrow] (hidden-\i-\j) -- (hidden-\nexti-\j);
    \fi

    \ifnum\i=5
      \draw[arrow] (hidden-\i-\j) -- (output-\j);
    \fi
  }
  \ifnum\i<6
    \foreach \j [evaluate=\j as \nextj using int(\j+1)] in {1,...,4} {
      \draw[arrow] (hidden-\i-\j) -- (hidden-\i-\nextj);
    }
  \fi
}
\end{tikzpicture}
\caption{A simplified representation of a decoder-only transformer model processing the input string \texttt{Mike Trout plays for the}. Each input token passes through several hidden layers. At each layer, each token is associated with a vector (represented by $\langle \textcolor{gray}{\tiny\bullet}, \textcolor{gray}{\tiny\bullet}, \textcolor{gray}{\tiny\bullet} \rangle$). The final hidden layer generates a unique probability distribution ($p_i$) over the next possible token for each input token.}
\label{fig:Transformer}
\end{figure}
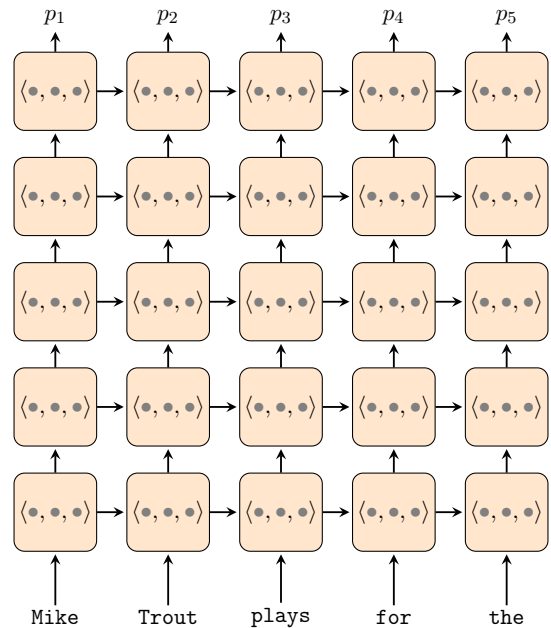

The language models we're interested in are transformer models \citep{VaswaniAttention}. In this section, we provide a basic understanding of how such models work.\footnote{For an in depth, conceptual overview of decoder only transformer models, see \citep{conceptual_guide_levinstein}.} In particular, we'll be focusing on autoregressive, decoder-only models such as OpenAI's GPT series and Meta's LLaMA series. The basic structure is as follows: 

\begin{enumerate}
    \item \textbf{Input Preparation:} Text data is fed to the model. For example, let's consider the phrase, \texttt{Mike Trout plays for the}. 

    \item \textbf{Tokenization:} The input text is tokenized, which involves breaking it down into smaller pieces called tokens. In English, these tokens are typically individual (sub)words or punctuation. So, our example sentence could be broken down into [\texttt{Mike}, \texttt{Trout}, \texttt{plays}, \texttt{for}, \texttt{the}].

    \item \textbf{Embedding:} Each token is then converted into a mathematical representation known as an embedding. This is a vector of a fixed length that represents the token along with its position in the sequence.\footnote{After the model is trained, intuitively what these embeddings are doing is representing semantic and other information about the token along with information about what has come before it in the sequence.} For instance, \texttt{Mike} might be represented by a list of numbers such as [0.1, 0.3, -0.2, \ldots].

    \item \textbf{Passing through Layers:} These embeddings are passed through a series of computational layers. Each layer transforms the embeddings of the tokens based on the each token's current embedding, as well as the information received from previous tokens' embeddings. This procedure enables information to be `moved around' from token to token across the layers. It is through these transformations that the model learns complex language patterns and relationships among the tokens. For example, to compute the embedding for \texttt{plays} in \texttt{Mike Trout plays for the} at a layer $m$, a decoder-only model can use information from the layer $m-1$ embeddings for \texttt{Mike}, \texttt{Trout}, and \texttt{plays}, but not from \texttt{for} or \texttt{the}. 

    \item \textbf{Prediction:} After the embeddings pass through the last layer of the model, a prediction for what the next token will be is made using the embedding \textit{just for} the previous token. This prediction involves estimating the probabilities of all potential next tokens in the vocabulary. When generating new text, the model uses this distribution to select the next token. For example, after processing the phrase \texttt{Mike Trout plays for the}, the model might predict \texttt{Angels} as the next token given its understanding of this sequence of text. (In reality, the model will actually make a prediction for what comes after each initial string of text. So, it will make predictions for the next token after \texttt{Mike}, after \texttt{Mike Trout}, after \texttt{Mike Trout plays}, etc.)
\end{enumerate}
The power of transformer models comes from their ability to consider and manipulate information across all tokens in the input, allowing them to generate human-like text and uncover deep patterns in language. \Cref{fig:Transformer} provides a basic depiction of information flow in decoder-only models. 

\section{Challenges in Deciphering the `Beliefs' of Language Models}
\label{challenges}

For now, let's assume that in order to generate human-like text, LLMs (like humans) have beliefs about the world. We might then ask how we can measure and discover their beliefs. This question immediately leads to a number of problems:

\subsection{Unreliable Self-Reporting} 
Asking an LLM directly about its beliefs is insufficient. As we've already discussed, models have a tendency to ``hallucinate'' or even lie. So belief reports alone cannot be taken as trustworthy. Moreover, when asked about its beliefs, an LLM likely will not ``introspect'' and decode some embedding that contains information about its information state. Instead, it just needs to answer the question in a reasonable way that accords with its training process. 

\subsection{Limited Behavioral Evidence}
When trying to understand human beliefs, we have a rich tapestry of behavioral evidence to draw upon. We consider not only what people say, but also what they do. For instance, if someone consistently invests in the S\&P, we infer that they believe the S\&P will go up in value, even if they never explicitly state it. For LLMs, however, we have a limited behavioral basis for inferring beliefs. The ``behavior'' of a language model is confined to generating sequences of tokens, which lacks the depth and breadth of human action.

\subsection{Contextuality of LLMs}
Everything one inputs and doesn't input into the LLM is fair game for it to base its responses on. Through clever prompting alone, there is no way to step ``outside'' of the language game the LLM is playing to get at what it \textit{really} thinks. This problem also plagues economists' and psychologists' attempts to uncover the beliefs of humans. For example, economists have challenged the validity of the famous ``framing effects'' of Tversky and Kahneman (\citeyear{tversky1981framing}) by considering the possibility that the subjects in the study updated on higher-order evidence contained in what was and wasn't said to them, and the rest of the context of the experiment (\cite{gilboa2020states}).\footnote{\cite{lieder2020resource} make a similar point.}

\subsection{Opaque and Alien Internal Structure}
While we can examine the embeddings, parameters, and activations within an LLM, the semantic significance of these elements is opaque. The model generates predictions using a complex algorithm that manipulates high-dimensional vectors in ways that don't obviously resemble human thought processes.

We can paraphrase a metaphor from Quine to help us think about language models:
\begin{quote}
    Different [models trained on] the same language are like different bushes trimmed and trained to take the shape of identical elephants. The anatomical details of twigs and branches will fulfill the elephantine form differently from bush to bush, but the overall outward results are alike. \cite[p. 7]{quine1960word}
\end{quote}
LLMs produce output similar to the output of humans competent in the same language. 
Transformer models are fundamentally different from humans in both structure and function. Therefore, we should exercise caution in interpreting their outputs and be aware of the inherent limitations in our understanding of their internal processes.

\section{Interpreting the Minds of LLMs}
\label{interpreting}

One potential strategy to decipher the beliefs of transformer models is to bypass the opacity of their internal structure using an approach known as ``probing'' \citep{alain2016understanding}. 

Although the internals of LLMs are difficult for humans to decipher directly, we can use machine learning techniques to create simplified models (probes) that can approximate or infer some aspects of the information captured within these internal structures.

At a high-level,  this works as follows. We generate true and false statements and feed them to the LLM. For each statement, we extract a specific embedding from a designated hidden layer to feed into the probe. The probe only has access to the embedding and is ignorant of the original text fed into the LLM. Its task is to infer the ``beliefs'' of the LLM solely based on the embedding it receives.

\begin{figure}
    \centering
    \begin{tikzpicture}[
  block/.style={rectangle, draw, fill=blue!20, 
    text width=9em, text centered, rounded corners, minimum height=4em},
  line/.style={draw, -latex}
]

\node [block] (input) {Generate True/False Statements};
\node [block, below=2cm of input] (llm) {LLM};
\node [block, below=2cm of llm] (llmout) {LLM Text Output};
\node [block, right=1cm of llm] (embed) {Embedding Extraction};
\node [block, right=1cm of embed] (probe) {Probe};
\node [block, below=2cm of probe] (beliefs) {Numerical Beliefs};

\path [line] (input) -- (llm);
\path [line] (llm) -- (llmout);
\path [line] (llm) -- (embed);
\path [line] (embed) -- (probe);
\path [line] (probe) -- (beliefs);

\end{tikzpicture}
    \caption{High-level overview of how the probe measures the beliefs of the LLM on inputs of true and false statements. Instead of looking at the text the LLM itself ouputs, we look at the numbers that the probe outputs.}
    \label{fig:overview}
\end{figure}
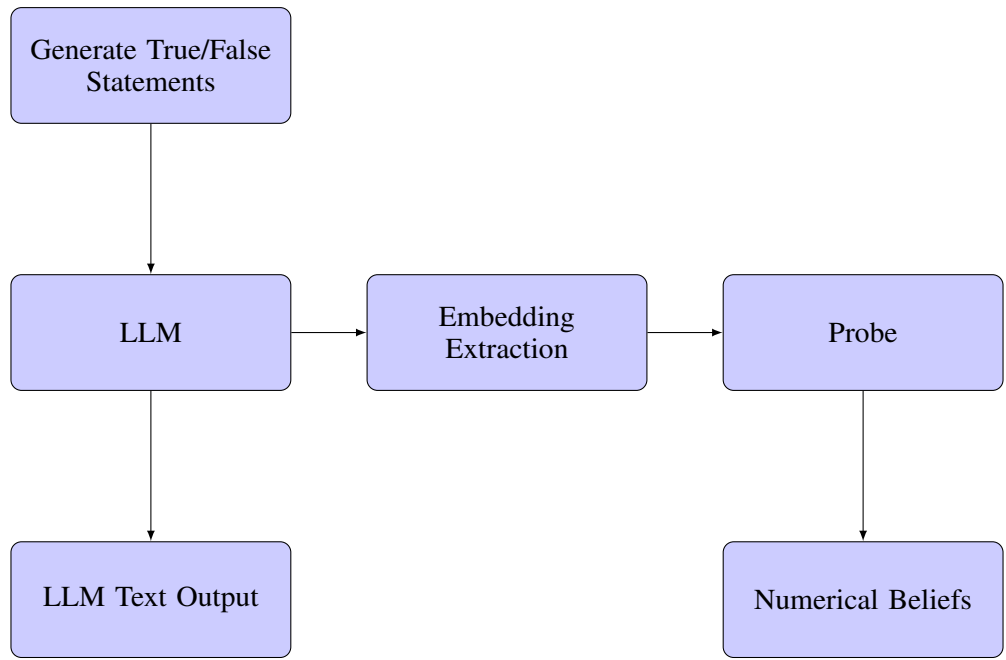

In practice, we focus on the embedding associated with the last token from a late layer. This is due to the fact that in autoregressive, decoder-only models like the LLMs we are studying, information flows forward. Therefore, if the LLM is processing a statement like \texttt{The earth is round}, the embeddings associated with the initial token \texttt{The} will not receive any information from the subsequent tokens. However, the embedding for the final word \texttt{round} has received information from all previous tokens. Thus, if the LLM computes and stores a judgement about the truth of the statement \texttt{The earth is round}, this information will be captured in the embedding associated with \texttt{round}.\footnote{The sentences in the dataset all ended with a period (i.e., full-stop) as the final token. We ran some initial tests to see if probes did better on the embedding for the period or for the penultimate token. We found it did not make much of a difference, so we did our full analysis using the embeddings for the penultimate tokens.} We use relatively late layers because it seems more likely that the LLM will try to determine whether a statement is true or false after first processing lower-level semantic and syntactic information in earlier layers.

\subsection{Supervised Learning Approach}

The first approach for training a probe employs supervised learning. This uses a list of statements labelled with their truth-values. The statements are each run through the language model. The probe receives as input the embedding for the last token from a specific layer of the large language model, and it outputs a number---intended to be thought of as a subjective probability---ranging from $0$ to $1$. The parameters of the probe are then adjusted based on the proximity of its output to the actual truth-value of the statement.

This approach was recently investigated by \cite{azaria2023internal}. They devised six labelled datasets, each named according to their titular subject matter: \texttt{Animals}, \texttt{Cities}, \texttt{Companies}, \texttt{Elements}, \texttt{Scientific Facts}, and \texttt{Inventions}. Each dataset contained a minimum of 876 entries, with an approximate balance of true and false statements, totalling 6,084 statements across all datasets. \Cref{tab:example_statements} provides some examples from these datasets.
\begin{center}
\begin{table}[h]
\centering
\caption{Example Statements from Different Datasets}
\label{tab:example_statements}
\begin{tabular}{@{}lll@{}}
\toprule
Dataset    & Statement & Label \\ 
\midrule
Animals & The giant anteater uses walking for locomotion. & 1 \\
        & The hyena has a freshwater habitat. & 0 \\
\midrule
Cities & Tripoli is a city in Libya. & 1 \\
       & Rome is the name of a country. & 0 \\
\midrule
Companies & The Bank of Montreal has headquarters in Canada. & 1 \\
          & Lowe's engages in the provision of telecommunication services. & 0 \\
\midrule
Elements & Scandium has the atomic number of 21. & 1 \\
         & Thalium appears in its standard state as liquid. & 0 \\
\midrule
Facts & Comets are icy celestial objects that orbit the sun. & 1 \\
      & The freezing point of water decreases as altitude increases. & 0 \\
\midrule
Inventions & Ernesto Blanco invented the electric wheelchair. & 1 \\
           & Alan Turing invented the power loom. & 0 \\
\bottomrule
\end{tabular}
\end{table}
\end{center}

\subsection{Azaria and Mitchell's Implementation}
\cite{azaria2023internal} trained probes on the embeddings derived from Facebook's OPT 6.7b model~\citep{zhang2022opt}.\footnote{The `6.7b' refers to the number of parameters (i.e., 6.7 billion).} Their probes were all feedforward neural networks comprising four fully connected layers, utilizing the ReLU activation function. The first three layers consisted of 256, 128, and 64 neurons, respectively, culminating in a final layer with a sigmoid output function. They applied the Adam optimizer for training, with no fine-tuning of hyperparameters, and executed training over five epochs.

For each of the six datasets, they trained three separate probes on the five other datasets and then tested them on the remaining one (e.g., if a probe was trained on \texttt{Cities}, \texttt{Companies}, \texttt{Elements}, \texttt{Facts}, and \texttt{Inventions}, it was tested on \texttt{Animals}). The performance of these probes was evaluated using binary classification accuracy. This process was repeated for five separate layers of the model, yielding fairly impressive accuracy results overall.

The purpose of testing the probes on a distinct dataset was to verify the probes' ability to identify a general representation of truth within the language model, irrespective of the subject matter.

\subsection{Our Reconstruction}
We implemented a reconstruction of Azaria and Mitchell's method with several modifications:

\begin{itemize}
    \item We constructed the probes for LLaMA 30b \citep{touvron2023llama}, a model from Meta with 33 billion parameters and 60 layers.
    \item We utilized an additional dataset named \texttt{Capitals} consisting of 10,000 examples, which was provided by Azaria and Mitchell. It has  substantial overlap with the \texttt{Cities} dataset, which explains some of the test accuracy.
    \item We trained probes on three specific layers: the last layer (layer -1), layer 56 (layer -4), and layer 52 (layer -8).
    \item We took the best of ten probes (by binary classification accuracy) for each dataset and each layer instead of the best of three.
\end{itemize}

Similar to the findings of Azaria and Mitchell, our reconstruction resulted in generally impressive performance as illustrated in \Cref{tab:classification_accuracy}.
\begin{table}
		\centering
		\begin{tabular}{lccccccc}  
			\toprule  
			& Animals & Capitals & Cities & Companies & Elements & Facts & Inventions \\  
			\midrule  
			Layer -1 & .722 & .970 & .867 & .722 & .755 & .826 & .781 \\  
			Layer -4 & .728 & .973 & .882 & .766 & .792 & .821 & .831 \\  
			Layer -8 & .729 & .967 & .869 & .742 & .694 & .810 & .792 \\  
			\bottomrule  
		\end{tabular}
        \vspace{1mm}
		\caption{Binary Classification Accuracy for probes trained on LLaMA 30b embeddings.}
      \label{tab:classification_accuracy}
\end{table}

\begin{figure}[h]
\centering
\includegraphics[width=\textwidth]{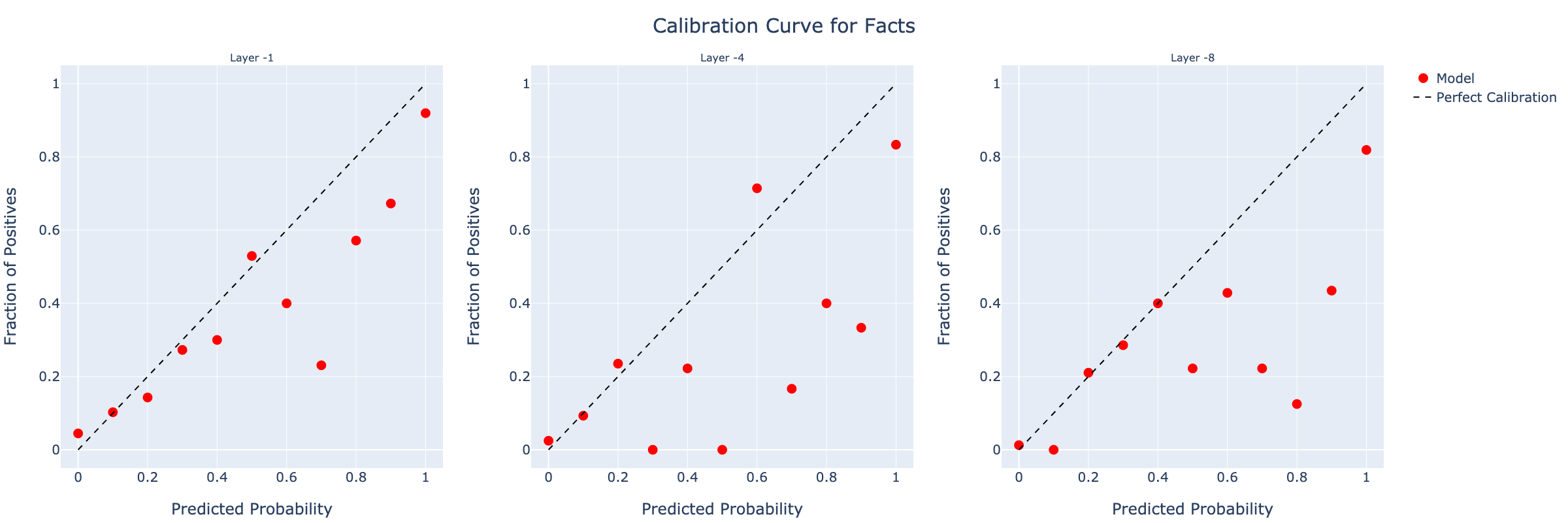}
\caption{Calibration curves for probes tested on the \texttt{Scientific Facts} dataset at each layer.}
\label{fig:calibration_subplot}
\end{figure}

In addition to binary classification accuracy, we evaluated the calibration of the probes across the different layers. Calibration provides another metric for evaluating the quality of the probes' forecasts. \Cref{fig:calibration_subplot} illustrates these calibration curves for each layer when tested on the \texttt{Scientific Facts} dataset.

\subsection{The Challenge of Generalization}

This section explores our empirical findings, which suggest that probes in this setting often learn features that correlate with truth in the training set, but do not necessarily generalize well to broader contexts.

\subsubsection{Evaluating Performance on Negations}

Creating Boolean combinations of existing statements is one of the most straightforward ways to generate novel statements for testing a model's generalization capabilities. Negation, the simplest form of Boolean operation, offers a useful starting point.\footnote{In formal models of beliefs and credence, the main domain is usually an algebra over events. If we wish to identify doxastic attitudes in language models, then we should check that those attitudes behave roughly as expected over such an algebra. Such algebras are closed under negation, so it is a motivated starting point.}

We derived \texttt{NegFacts} and \texttt{NegCompanies} from Azaria and Mitchell's datasets. These new datasets contained the negations of some statements in \texttt{Scientific Facts} and \texttt{Companies} respectively. For instance, the statement \texttt{The earth orbits the sun} from \texttt{Scientific Facts} is transformed into \texttt{The earth doesn't orbit the sun} in \texttt{NegFacts}.

Given that the original datasets contained few Boolean statements, these negation datasets allowed us to test the probes on a simple new distribution.


We initially tested the probes trained on \Animals, \Capitals, \Cities, \Companies, \Elements, and \Inventions~ (i.e., trained all positive datasets except \texttt{Scientific Facts}) on \texttt{NegFacts}. Similarly, we tested the probes trained on \Animals, \Capitals, \Facts, \Cities, \Elements, and \Inventions on \NegCompanies. Since roughly 50\% of the statements in each of \texttt{NegFacts} and \NegCompanies are true, the accuracy of five of six of these probes was worse than chance, as \Cref{tab:negfacts_performance} illustrates.

We then tested a new set of probes on \NegFacts, after training on all seven original datasets (including \Facts) and \NegCompanies, which consisted of 550 labeled negations of statements from \Companies. Thus, these probes were trained on \textit{all positive variants of the negated statements they were tested on, along with all positive examples from \Companies~ and their negated counterparts.} We did the same, \textit{mutatis mutandis} with \NegCompanies. Despite the expanded training data, the performance was still surprisingly poor, as shown in \Cref{tab:negfacts_performance}.

\begin{table}[h!]
\centering
\caption{Binary classification accuracy for NegFacts compared to Facts. `NegFacts\textsuperscript{1}' (`NegCompanies\textsuperscript{1}') denotes the accuracy for probes trained only on positive datasets, excluding \Facts~ (\Companies). `NegFacts\textsuperscript{2}' denotes the accuracy for probes trained on all positive datasets including \Facts~ and \NegCompanies, while `NegCompanies\textsuperscript{2} denotes the accuracy for probes trained on all positive datasets including \Companies and \NegFacts.}
\begin{tabular}{lccccccc}
\toprule
Layer & Facts & NegFacts\textsuperscript{1} & NegFacts\textsuperscript{2} &Companies & NegCompanies\textsuperscript{1} & NegCompanies\textsuperscript{2}\\
\midrule
-1 & .826 & .408 & .526 & .722 & .555 & .567\\
-4 & .821 & .373 & .568 & .766 & .460 & .629 \\
-8 & .810 & .373 & .601 & .742 & .431 & .596\\
\bottomrule
\end{tabular}
\label{tab:negfacts_performance}
\end{table}
Since the probes failed to do well on \NegFacts and \NegCompanies even after training on all positive analogs along with other negative examples, it's likely the original probes are not finding representations of truth within the language model embeddings. Instead, it seems they're learning some other feature that correlates well with truth on the training sets but that does not correlate with truth in even mildly more general contexts. 

Of course, we could expand the training data to include more examples of negation and other Boolean combinations of sentences. This likely would allow us to train better probes. However, we have general conceptual worries about generalizing probes trained with supervised learning that we will explore in the next subsection. Specifically, we will be delving into the potential shortcomings of relying on supervised learning techniques for probe training. These issues stem from the inherent limitations of supervised learning models and how they handle unknown scenarios and unseen data patterns.

\subsection{Conceptual Problems: Failure to Generalize}
\label{concept:general}

In the realm of machine learning, out-of-distribution generalization remains a pervasive challenge for classifiers. One of the common pitfalls involves learning \textit{spurious correlations} that may be present in the training data, but do not consistently hold in more general contexts.

Consider an example where a classifier is trained to distinguish between images of cows and camels \citep{beery2018recognition}. If the training set exclusively features images of cows in grassy environments and camels in sandy environments, the classifier may learn to associate the environmental context (grass or sand) with the animal, and using that to predict the label, rather than learning the distinguishing features of the animals themselves. Consequently, when presented with an image of a cow standing on sand, the classifier might erroneously label it as a camel. 

We think there are special reasons to be concerned about generalization when training probes to identify a representation of truth using supervised learning because supervised learning severely limits the sort of data we can use for training and testing our probes. First, we need to use sentences we believe the model itself is in a position to know or infer from its own training data. This is the easier part. The harder part is curating data that we can unambiguously label correctly.  The probe most directly is learning to predict the \textit{label}, not the actual truth-value. These coincide only when the labels are completely correct about the statements in the training and test set. 

We ultimately want to be able to use probes we've trained on sentences whose truth-value we ourselves don't know. However, the requirement that we accurately label training and testing data limits the confidence we can place in the probes' capability of accurately identifying a representation of truth within the model. For instance, consider the following statements:
\begin{itemize}
    \item Barry Bonds is the best baseball player of all time.
    \item If the minimum wage is raised to \$15 an hour, unemployment will increase. 
    \item France is hexagonal.
    \item We are morally responsible for our choices.
    \item Caeser invaded Gaul due to his ambition. 
\end{itemize}
These statements are debatable or ambiguous. We must also be cautious of any contentious scientific statements that lack full consensus or could be reconsidered as our understanding of the world evolves.

Given these restrictions, it's likely the probes will  identify properties that completely or nearly coincide with truth over the limited datasets used for training and testing. For instance, the probe might identify a representation for:
\begin{itemize}
    \item Sentence is true \textit{and} contains no negation
    \item Sentence is true \textit{and} is expressed in the style of Wikipedia
    \item Sentence is true \textit{and} can be easily verified online
    \item Sentence is true \textit{and} verifiable
    \item Sentence is true \textit{and} socially acceptable to assert
    \item Sentence is true \textit{and} commonly believed
    \item Sentence is true \textit{or} asserted in textbooks
    \item Sentence is true \textit{or} believed by most Westerners
    \item Sentence is true \textit{or} ambiguous
    \item Sentence is accepted by the scientific community
    \item Sentence is believed by person X
\end{itemize}


On the original datasets we used, if the probe identified representations corresponding to any of the above, it would achieve impressive performance on the test set. Although we can refine our training sets to eliminate some of these options, we won't be able to eliminate all of them without compromising our ability to label sentences correctly. 

Indeed, if the labels are inaccurate, the probe might do even better if it identified properties like ``Sentence is commonly believed'' or ``Sentence corresponds to information found in many textbooks'' even when the sentence is not true.\footnote{
    \cite{azaria2023internal} did an admirable job creating their datasets. Some of the statements were generated automatically using reliable tables of information, and other parts were automated using ChatGPT and then manually curated. Nonetheless, there are some imperfect examples. For instance, in \texttt{Scientific Facts}, one finds sentences like \texttt{Humans have five senses: sight, smell, hearing, taste, and touch}, which is not unambiguously true. 
}

This situation can be likened to the familiar camel/cow problem in machine learning. But given the constraints imposed by using supervised learning and limited data, isolating representations of truth from other coincidental properties is even more challenging than usual. The fact that probes empirically seem to identify representations of something other than truth should make us wary of this method. 

\subsection{Conceptual Problems: Probabilities Might not Correspond to Credences}

So far we have been assuming that if the probes extracted accurate probabilities, that this would be good evidence we were extracting the credences of the model. However, this is too quick. While these probes output probabilities for statements, these probabilities do not directly correspond to the ``credences'' of the underlying language model. This disparity arises because the \textit{probe} is directly penalized based on the probabilities it reports, while the underlying model is not. Thus, the probe aims to translate the information embedded within the language model's representations into probabilities in a manner that minimizes its own loss.

Consider an illustrative analogy: Suppose I forecast stock prices, with rewards based on the accuracy of my predictions. However, my predictions are entirely reliant on advice from my uncle, who, unfortunately, is systematically inaccurate. If he predicts a stock's price will rise, it actually falls, and vice versa. If I wish to make accurate forecasts, I need to reverse my uncle's predictions. So, while my predictions are based entirely on my uncle's advice, they don't directly reflect his actual views. Analogously, the probe makes predictions based on the information in the embeddings, but these predictions don't necessarily represent the actual ``beliefs'' of the language model.

This analysis suggests that there are further conditions that probabilities extracted by a probe must satisfy in order to be properly considered credences. Going back to the example of my uncle, the problem there was that the predictions I was making were not used by my uncle in the appropriate way in order to make decisions. Thus it makes sense to day that my predictions do not reflect my \textit{uncle's} views. 

Thus, beyond merely extracting probabilities that minimize loss, there are \textit{other} requirements that extracted representations must satisfy. In the context of natural langauge processing, \citeauthor{harding2023operationalising} in a recent paper (\citeyear{harding2023operationalising}) has argued for three conditions that must hold of a a pattern of activations in neural models for it to count as a representation of a property:\footnote{\citeauthor{harding2023operationalising} makes these conditions precise in the language of information theory. Further development of the concept of representation in the context of probes strikes us as an important line of research in working to understand the internal workings of deep learning models.}
\begin{enumerate}
    \item \textbf{Information.} The pattern must have information about the property.
    \item \textbf{Use.} The system (in our case, LLM) must use the pattern to accomplish its task.
    \item \textbf{Misrepresentation.} It should be possible for the pattern to misrepresent the property.
\end{enumerate}

In our context the property we care about is \textit{truth}, and we might call the corresponding representation \textit{belief}. Thus we see what went wrong in the uncle example: even though the predictions I extracted from his advice were informative (satisfied \textit{information}), they also violated \textit{use}: my uncle did not use them, but instead used his own forecasts (benighted as they were) to buy stocks. So it didn't make sense to refer to \textit{my} forecasts as \textit{his} beliefs. For this same reason, depending on how the actual LLM ends up using the representation the probe extracts, it might not make sense to call the \textit{probe's} outputs the \textit{LLM's} beliefs.

\subsection{Unsupervised Learning: CCS}
The second approach for training a probe eschews the need for labelled data. Instead, it attempts to identify patterns in the language model's embeddings that satisfy certain logical coherence properties.

One particularly innovative implementation of this idea is the Contrast-Consistent Search (CCS) method proposed by \cite{burns2022discovering}. The CCS method relies on training probes using \textit{contrast pairs}. For our purposes, we can think of a contrast pair as a set of statements $x^+$ and $x^-$, where $x^+$ has no negation, and $x^-$ is the negated version of $x^+$. For example, \texttt{The earth is flat} and \texttt{The earth is not flat} form a contrast pair. (One can also form contrast pairs picking up on other features instead. For example, \cite{burns2022discovering} uses movie reviews from the IMDb database \citep{maas2011learning} prefixed with ``The following movie review expresses a positive sentiment'' and ``The following move review expresses a negative sentiment'' to create contrast pairs.)

CCS proceeds in the following manner:
\begin{enumerate}
    \item Create a dataset of contrast pairs of true or false statements. Each pair is of the form $(x_i^+, x_i^-)$, so the dataset is 
    $\{(x_1^+, x_1^-), \ldots, (x_n^+, x_n^-)\}$.
    \item Pass each statement through the network, and extract the embedding for the last token from a chosen layer. 
    \item Train a probe $p_\theta$ with parameters $\theta$. The probe takes these embeddings as inputs and outputs numbers between $0$ and $1$. It is trained such that:
    \begin{enumerate}
        \item The probabilities given by the probe for the embeddings of $x_i^+$ and $x_i^-$ should sum up to (approximately) 1.
        \item The probabilities given by the probe for the embeddings of $x_i^+$ and $x_i^-$ are distinct. 
    \end{enumerate}
\end{enumerate}
The underlying rationale behind step 3(a) is that if the model represents $x_i^+$ as true, then it should represent $x_i^-$ as false and vice versa. We can think of a successful probe as encoding a probability function (or something approximating a probability function) that underwrites the beliefs of the model. Thus, if a probe is able to find this representation within the embeddings, it should map the embeddings of $x_i^+$ and $x_i^-$ to numbers whose sum is close to 1. This is the central insight behind \citeauthor{burns2022discovering}'s approach. As they put it, CCS finds a ``direction in activation space that is consistent across negations'' (p. 3). Step 3(b) is crucial in preventing the probe from trivially mapping every embedding to $.5$ to satisfy condition 3(a).

To implement the conditions in step 3, \cite{burns2022discovering} introduce two loss functions. The consistency loss, given by
$$L_\text{consistency}(\theta; x_i) \coloneqq (1 - p_\theta (\textrm{emb}(x_i^+)) - p_\theta (\textrm{emb}(x_i^-)))^2,$$
penalizes a probe for mapping the embeddings for $x_i^+$ and $x_i^-$ to numbers whose sum deviates from $1$. (Here $\textrm{emb}(x)$ denotes the embedding for $x$'s last token at the given layer.)

The confidence loss, defined as
$$L_\text{confidence}(\theta; x_i) \coloneqq \min \{p_\theta (\textrm{emb}(x_i^+)), p_\theta (\textrm{emb}(x_i^-))\}^2,$$
penalizes a probe for approximating the degenerate solution of returning $.5$ for every embedding.\footnote{
    Some readers may worry about a second degenerate solution. The model could use the embeddings to find which of $x_i^+$ and $x_i^-$ contained a negation. It could map one of the embeddings to (approximately) $1$ and the other to (approximately) $0$ to achieve a low loss. \cite{burns2022discovering} avoid this solution by normalizing the embeddings for each class by subtracting the means and dividing by the standard deviations. However, as we'll see below, for the datasets that we used, such normalization was ineffective, and the probes consistently found exactly this degenerate solution. 
} 

The total loss for the dataset, termed the CCS loss, is given by:
$$L_\text{CCS}(\theta) \coloneqq \frac{1}{n}\sum_{i=1}^n L_\text{consistency}(\theta; x_i) + L_\text{confidence} (\theta; x_i). $$
Crucially, this loss function does not take actual accuracy into account. It merely penalizes probes for lack of confidence and (one type of) probabilistic incoherence. 

An important caveat to note is that, while the trained CCS probe itself approximates probabilistic coherence, its outputs do not correspond to the credences or subjective probabilities of the model. $L_\text{confidence}$ pushes the probe to report values close to $0$ or $1$ only. To see why, suppose a probe at one stage of the training process returned $.6$ for $x_i^+$ and $.4$ for $x_i^-$. It could get a better loss by reporting $.99$ for $x_i^+$ and $.01$ for $x_i^-$ regardless of the language model's actual subjective probabilities, and it will be pushed in this extreme direction by gradient descent. So, the probes themselves are, at best, useful for determining what the model's categorical beliefs are, not its probabilities.\footnote{
    One way to see that $L_\text{CCS}$ won't incentive a probe to learn the actual credences of the model is to observe that this loss function is not a strictly proper scoring rule \citep{gneiting2007strictly}. However, use of a strictly proper scoring rule for training probes requires appeal to actual truth-values, which in turn requires supervised learning. 
}

\cite{burns2022discovering} report two key findings. First, even when using a fully linear probe, CCS yields high accuracy rates---often over 80\%---across numerous datasets for a number of different language models.\footnote{A linear probe is one that applies linear weights to the embeddings (and perhaps adds a constant), followed by a sigmoid function to turn the result into a value between $0$ and $1$. Linear probes have an especially simple functional form, so intuitively, if a linear probe is successful, the embedding is easy to extract.} Second, binary classification using CCS tends to be slightly more accurate than the LLMs' actual outputs when asked whether a statement is true. This suggests that CCS can identify instances where the language models internally represent a statement as true but output text indicating it as false, or vice versa. (For a detailed description of their results, see p. 5 of their paper).

However, the performance of the CCS probe on GPT-J \citep{gpt-j}, the only decoder-only model tested in the study, was less impressive, with an accuracy rate of only 62.1\% across all datasets. This is notably lower than the peak accuracy of 84.8\% achieved by the encoder-decoder model UnifiedQA \citep{khashabi2020unifiedqa}.

\subsection{Our Reconstruction}
We reconstructed Burns et al.'s method using embeddings for LLaMA 30b with probes trained and tested on contrast pairs from the \texttt{Scientific Facts} and \texttt{NegFacts} datasets, as well as the \texttt{Companies} and \texttt{NegCompanies} datasets. These contrast pairs consist of simple sentences and their negations. This approach more closely resembles the examples given in the main text of Burns et al.'s paper, than do the longer and more structured contrast pairs that they actually used to train their probes, such as movie reviews from IMDb.

We experimented with a variety of different methods and hyperparameters. However, we found that while CCS probes were consistently able to achieve low loss according to $L_\text{CCS}$, their accuracy was in effect no better than chance---it ranged from 50\% to 57\% depending on the training run. (Recall, the minimum possible accuracy for a CCS probe is 50\%.) Low accuracy persisted even after we normalized the embeddings for each class by subtracting the means and dividing by the standard deviations, following the same procedure as \cite{burns2022discovering}.

Upon inspection, it is clear that CCS probes were usually able to achieve low loss simply by learning which embeddings corresponded to sentences with negations, although they sometimes learned other features uncorrelated with truth. Given the similarity of the outcomes across these experiments, we report quantitative results from the probes we trained using a simple one hidden layer MLP with 100 neurons followed by a sigmoid output function on layers 60, 56, and 52 in \Cref{tab:CCS_performance}. Recall these layers correspond to the last, fourth-last, and eighth-last layers of the LLaMA 30b, respectively.

\begin{table}
		\centering
		\begin{tabular}{lcccc}  
			\toprule  
			Layer & $L_\text{CCS}$ & $L_\text{Confidence}$ & $L_\text{Consistency}$ & Accuracy \\  
			\midrule  
			 -1 & .009 & .004 & .005 & .552 \\  
			 -4 & .003 & .002 & .002 & .568 \\  
			 -8 & .013 & .002 & .010 & .502 \\  
			\bottomrule  
		\end{tabular}
        \vspace{1mm}
		\caption{Performance of CCS Probes at various layers on each component of the loss function and in terms of overall accuracy.}
        \label{tab:CCS_performance}
\end{table}

We can confirm that, despite normalization, the probes were able to determine which embeddings corresponded to positive and negative examples in layers -1 and -4 by checking the average values the probes returned for members of each class. Probes found some other way to achieve low loss in layer -8, but they did not do any better in terms of accuracy as shown in \Cref{tab:CCS_confidence}. (Recall, only roughly half the positive examples and half the negative examples are actually true.)

\begin{table}
		\centering
		\begin{tabular}{lcc}  
			\toprule  
			Layer & Positive Prediction Avg & Negative Prediction Avg \\  
			\midrule  
			 -1 & .968 & .035 \\  
			 -4 & .990 & .012 \\  
			 -8 & .389 & .601 \\  
			\bottomrule  
		\end{tabular}
        \vspace{1mm}
		\caption{Average prediction value in positive examples and negative examples at each layer.}
        \label{tab:CCS_confidence}
\end{table}

Now, one might think that this failure of our probes is itself fragile. Normalization by subtracting the mean and dividing by the standard deviation was supposed to disguise the grammatical form of the sentences, but it did not. There is likely some more sophisticated normalization method that would work better. 

We agree that such alternative methods are likely possible. However, as we discuss in the next section, we are not sanguine about the basic approach \cite{burns2022discovering} use for conceptual reasons.

\subsection{Conceptual Problems: Failure to Isolate Truth}
The advantage of CCS and unsupervised approaches more generally over supervised approaches is that they do not restrict the training and testing data so severely. There is no need to find large collections of sentences that can unambiguously be labeled as true or false. So, one may have hope that CCS (and unsupervised approaches) will generalize well to new sentences because we are less restricted in training.

However, the fundamental issue we've identified is that coherence properties alone can't guarantee identification of truth. As demonstrated in our experiments, probes might identify sentence properties, such as the presence or absence of negation, rather than truthfulness.

Further, probes could identify other, non-truth-related properties of sentences. For example, they could associate truth with widespread belief, resulting in the classification ``$x$ is true \textit{and} commonly believed'' or even ``$x$ is believed by most people''.

To demonstrate this, consider any probability function $\Pr$. The sum of the probabilities that a sentence $x$ is true and commonly believed, and that it is false or not commonly believed, equals 1. Indeed, this equation holds for any sentence property $P$, where $\Pr(x \wedge P(x)) + \Pr(\neg x \vee \neg P(x))=1$. Likewise, $\Pr(x \vee P(x)) + \Pr(\neg x \wedge \neg P(x)) =1$.\footnote{These are both consequences of the fact that for any proposition $A$, $\Pr(A) + \Pr(\neg A) = 1$: take $A \coloneqq x \wedge \Pr(x)$, for example, and apply de Morgan's laws.} Checking for coherence over all Kolmogorov probability axioms---which require probabilities to be non-negative, normalized, and additive---will rule out some properties $P$, but will not come close to isolating truth. This means that coherence criteria alone can't distinguish encodings of truth from encodings of other concepts.

The failure to isolate truth here is reminiscent of the issue we noted with supervised learning, where truth may align with some alternative property over a dataset. However, the reasons for the failure differ. In the case of CCS and other unsupervised methods, the problem lies in the inability of formal coherence patterns alone to separate the encoding of truth from the encoding of other properties that differentiate positive from negative examples. If it's generally easier to find ``directions in activation space'' that differentiate examples but don't correspond exclusively to truth, then CCS probes will either fail immediately or fail to generalize.\footnote{\cite{burns2022discovering} investigate other unsupervised approaches as well that appeal to principal component analysis and/or clustering (such as Bimodal Salience Search (p. 22)). We believe---with some changes---most of the conceptual issues for CCS apply to those as well.}


\section{Do LLMs even have beliefs at all?}
\label{belliefsatall}

Our investigation points in a negative direction: probing the beliefs of LLMs is more difficult than it appeared after a first pass. Does this mean that we should be skeptical that LLMs have beliefs all together? 

To gain traction on this question we will consider arguments that intend to show that LLMs cannot have beliefs, even in principle. These arguments rely on the claim that LLMs make predictions about which tokens follow other tokens, and do not work with anything like propositions or world-models.

We claim that these arguments are misguided. We will show that our best theories of belief and decision making make it a very live possibility that LLMs \textit{do} have beliefs, since beliefs might very well be helpful for making good predictions about tokens. We will argue that ultimately whether or not LLMs have beliefs is largely an empirical question, which motivates the development of better probing techniques. 

\subsection{Stochastic Parrots \& the Utility of Belief}
\label{stocpar}

Even without having known the limitations of current probing techniques, some have expressed deep skepticism that LLMs have anything resembling beliefs. For example, \cite{bender2021dangers} write: 

\begin{quote}
Text generated by an LM is not grounded in communicative intent, any model of the world, or any model of the reader's state of mind. It can't have been because the training data never included sharing thoughts with a listener, nor does the machine have the ability to do that\ldots an LM is a system for haphazardly stitching together sequences of linguistic forms it has observed in its vast training data, according to probabilistic information about how they combine, but without any reference to meaning: a stochastic parrot. (pp. 616-617)
\end{quote}

Similarly, Shanahan (\citeyear{shanahan2022talking}) writes, 

\begin{quote}
    A bare-bones LLM doesn’t “really” know anything because all it does, at a fundamental level, is sequence prediction. Sometimes a predicted sequence takes the form of a proposition. But the special relationship propositional sequences have to truth is apparent only to the humans who are asking questions\ldots Sequences of words with a propositional form are not special to the model itself in the way they are to us. The model itself has no notion of truth or falsehood, properly speaking, because it lacks the means to exercise these concepts in anything like the way we do. (p. 5)
\end{quote}

These arguments rely on the idea that all the LLM is doing is predicting the next token. Because of this, both deny that the LLM can be working with anything like a meaningful model of the world. In other words, there is nothing \textit{propositional} going on under the hood.   

Shanahan doesn't deny that LLMs might contain information about the world around them. He does, however, claim that LLMs don't make judgements or have beliefs: 

\begin{quote}
    Only in the context of a capacity to distinguish truth from falsehood can we legitimately speak of “belief” in its fullest sense. But an LLM — the bare-bones model — is not in the business of making judgements. It just models what words are likely to follow from what other words. The internal mechanisms it uses to do this, whatever they are, cannot in themselves be sensitive to the truth or otherwise of the word sequences it predicts. Of course, it is perfectly acceptable to say that an LLM ``encodes”, ``stores”, or ``contains” knowledge, in the same sense that an encyclopedia can be said to encode, store, or contain knowledge\ldots But if Alice were to remark that ``Wikipedia knew that Burundi was south of Rwanda'', it would be a figure of speech, not a literal statement. (p. 5)
\end{quote}

The idea is that, since the LLM models which tokens are likely to follow other tokens, and doesn't interact with the world in any other way, it cannot be tracking the truth. This is similar to the argument in the \citeauthor{bender2021dangers} quote above: since the LLM does not have ``communicative intent'', it cannot be using any model of the world or the reader to make its predictions.

These arguments, however, rest on a mistake. While it is true that the ultimate output of an LLM is a token sampled from a probability distribution over tokens, and so the LLM is certainly modeling what words are probable to come after other words, this does \textit{not} mean that the internal mechanisms must be insensitive to truth. This is because it might very well \textit{be} that a capacity to distinguish truth from falsehood is very useful for predicting the next token. In other words, tracking the truth of propositions could be a good \textit{means} toward the end of predicting what token comes next.

This is in line with  a much more general feature of many types of goal directed action that can be made precise with decision theory. Decision theory gives us our best models of rational choice. The core idea of decision theory is an \textit{expected utility maximizer}. When faced with a set of options, an expected utility maximizer combines two different attitudes to compute which act to take: beliefs in the form of a probability function, and desires in the form of a utility function.\footnote{The canonical formalization of this idea in economics and statistics is Savage's \textit{Foundations of Statistics} (\citeyear{savage1972foundations}). Philosophers use Savage's formulation, as well as Jeffrey's in \textit{The Logic of Decision} (\citeyear{jeffrey1990logic}).} There is a precise sense in which all the agent cares about is the \textit{utility}.\footnote{More precisely, utility is a numerical representation that captures how strongly an agent cares about outcomes.} The agent does not care about belief for its own sake, but does have beliefs in order to take effective action. 

For example, an investor may care purely about the return on her investments. She may take actions with the goal to maximize her profit. It would be a mistake to conclude from this that the investor must not have beliefs, because she is merely doing profit maximization. Indeed, the investor's beliefs about how various firms will perform will probably play a crucial role in helping her make decisions. 

Similarly, it is a mistake to infer from the fact that the LLM outputs tokens that are likely to follows its inputs that the LLM must not have beliefs. On the contrary, given that our best theories of intelligent behaviour involve belief as a crucial component, it should be a very live hypothesis that the LLM is doing its best to track truths about the world, \textit{in order to} maximize predictive accuracy.\footnote{We are here ignoring nuances involving inner alignment \citep{hubinger2019risks}.}

Even beyond decision theory, philosophers have long held that true beliefs are useful for achieving goals, and that they play the functional role of helping us take successful action (\cite{millikan1995white}; \cite{papineau1988reality}). Indeed, not only is it useful, but it is a common view that the instrumental utility of accurate beliefs applies selection pressure on agents and organisms to conform to epistemic norms (\cite{street2009evolution}; \cite{cowie2014defence}). For example, in the context of forming true beliefs by induction, Quine famously writes, ``[c]reatures inveterately wrong in their inductions have a pathetic but praiseworthy tendency to die before reproducing their kind'' (p. 13, \citeyear{quine1969natural}).

This is very intuitive. It is easy to generate decision contexts (such as strategic board games, investing, figuring out how to get to Toronto from Prague, etc.) that do seem to push us to form accurate beliefs about the world.

This is not to say that it is \textit{necessary} that LLMs have beliefs, or that they necessarily have accurate beliefs. There are contexts where there seems to be less pressure on us to form accurate beliefs (\cite{stich1990fragmentation}). Importantly, there are two sub-cases to consider here. The first is the case in which there is little or no selection pressure for forming true beliefs, but there is not selection against \textit{having} beliefs. For example, \cite{smead2009social} considers contexts in which there are evolutionary advantages for misperceiving the payoffs of a strategic interaction (section 3.4). The second is the one in which there is selection pressure against having beliefs altogether (or, more conservatively, there is no selection pressure for having beliefs). For example, \citeauthor{godfrey1991signal} (\citeyear{godfrey1991signal}, \citeyear{godfrey1998complexity}), \cite{sober1994adaptive}, \cite{stephens2001selectively}, and \cite{smead2015role} have all developed models that characterize when an agent should (be expected to) learn from its environment and then select actions based on what it learned, and when it should not. This later situation is one in which there is selection pressure against (or at least none for) forming beliefs.

This leads us to the conclusion that, whether or not LLMs have beliefs, is largely an empirical matter. In favour of the view expressed by folks like \citeauthor{shanahan2022talking} and \citeauthor{bender2021dangers}, there certainly are contexts in which there is little to no selection pressure in favour of accurate beliefs, and indeed there are contexts that push against having beliefs altogether. On the other hand, there are plenty of contexts in which it is very useful to have an accurate map of the world, in order to guide action. Indeed, out best theories of rational choice witness this.

\section{Probing the Future}
\label{con}

We've deployed both empirical and conceptual tools in order to investigate whether or not LLMs have beliefs and, if so, how we might measure them. We demonstrated that current probing techniques fail to generalize adequately, even when we set aside the conceptual question of whether or it not it makes sense to ascribe beliefs to LLMs in the first place. We then considered two prominent arguments against the claim that LLMs have beliefs, and showed that they rest on a mistake. Ultimately, the status of beliefs in LLMs is (largely) an empirical question. 

Here we outline two possible empirical paths one might pursue to make progress on measuring beliefs. These are certainly not the only two. Finally, we discuss a possible line of investigation one might take in order gain clarity on whether or not it makes sense to attribute beliefs to LLMs. 

\subsection{Applying Pressure for Truth}

One of the insights in \cref{stocpar} is that contexts can apply different amounts of pressure to track the truth. This suggests a natural experiment to run: use prompt engineering to apply more pressure for the LLM to track truth, and then probe the LLM. For example, at a first pass, instead of just inputting a sentence like, \texttt{Rome is the name of a country}, we might input the prompt, \texttt{I want you to think hard about the following sentence: ``Rome is the name of a country.''} Perhaps refined versions of such  prompting will make the representations of truth more perspicuous and easier for probes to find.\footnote{The conceptual problems plaguing both types of probing techniques would still exist. However, the thought is that if the representation of truth is especially perspicuous, the probes might land upon such a representation naturally. This is a highly empirical question.}

\subsection{Replacing Truth with Chance}

So far the prompts we tested the probes on all had clear truth values. This has the advantage that we can use them for supervised learning, since we have the labels (the truth values). However, it also limits the ability to generate large datasets, since generating and labeling true sentences is costly. Furthermore, if we are selecting sentences that we think it is plausible the LLM might ``know'' (have high credence in), then our probing technique might not get good feedback on what the state of the LLM is like when it is more uncertain. 

One way we might address these issues is by systematically generating prompts that describe chance set-ups, and then training and testing the probe on statements about outcomes. For example, we might prompt the model with something like, \texttt{There is an urn with six yellow balls, four purple balls, and no\-thing else. A ball is drawn uniformly at random.} Then, within the scope of that prompt, we can use statements like \texttt{The ball drawn is purple}, which we can easily label with a chance value of 0.4.

\subsection{The Question of World-Models}

In \cref{stocpar} we pushed back against the claims in \cite{shanahan2022talking} and \cite{bender2021dangers} that all an LLM is doing is text prediction, and thus cannot be tracking the truth. We did this by showing that there are many contexts in which tracking the truth is useful for other goals. We did not, however, fully address other parts of their arguments. In particular, both \citeauthor{shanahan2022talking} and \citeauthor{bender2021dangers} are worried that the predictions of LLMs are not generated using any ``model of the world'' (p. 616, \cite{bender2021dangers}), and thus that the internal states cannot ``count as a belief about the world'' (p. 6, \cite{shanahan2022talking}).

In contrast to the concern about doing mere sequence prediction, this concern focuses our attention on \textit{how} the LLM computes the distribution, and whether or not it does so in a way that corresponds to it having any kind of model of the world. For example, even if there is some sense in which the LLM does keep track of something like truth because of pressure to do so as described in \cref{stocpar}, a skeptic might reply that the representation the system is using is not truth-apt. 

We thus think that a productive way to proceed would be to characterize the \textit{latent variables}\footnote{Latent variables are best understood in contrast to \textit{observable} variables. Suppose, for example, that you are trying to predict the outcomes of a series of coin tosses. The observable variables in this context would be the actual outcomes: heads and tails. A latent variable would be an \textit{unobservable} that you use to help make your predictions. For example, suppose you have different hypotheses about the bias of the coin and take the expected bias as your prediction for the probability of heads on the next toss. You are using your beliefs about the latent variable (the bias of the coin) to generate your beliefs about the observables.} that an LLM uses to predict text, and to evaluate whether or not these variables correspond to anything we might want to consider a world-model.\footnote{Using latent variables to compute probability distributions is commonplace in science and statistics (\cite{everett2013introduction}). Though we do not have the space to do latent variable methods full justice, one reason for this is that using distributions over latent variables in order to calculate a distribution over observable variables can have massive computational benefits (see, for example, chapter 16 of \cite{goodfellow2016deep}). Thus, it would be fairly surprising if there \textit{weren't} a useful way to think of LLMs as using some kinds of latent variables in order to make predictions about the next token. Indeed, there is already some preliminary work on what sorts of latent variables LLMs might be working with (\cite{xie2021explanation}; \cite{jiang2023latent}).} This requires both empirical work (in order to understand what kind of latent variables LLMs actually work with) and conceptual work (in order to understand what it would take for a latent variable to be truth-apt).\footnote{This question is related to the classic \textit{theoretician's dilemma} in the philosophy of science (\cite{Hempel1958-HEMTTD}).}

\section*{Acknowledgments}
Thanks to Amos Azaria, Dylan Bowman, Nick Cohen, Jacqueline Harding, Aydin Mohseni, Bruce Rushing, Nate Sharadin, and audiences at UMass Amherst and the Center for AI Safety for helpful comments and feedback. Special thanks to Amos Azaria and Tom Mitchell jointly for access to their code and datasets. We are grateful to the Center for AI Safety for use of their compute cluster. B.L. was partly supported by a Mellon New Directions Fellowship (number 1905-06835) and by Open Philanthropy. D.H. was partly supported by a Long-Term Future Fund grant.

\bibliographystyle{chicago}
\bibliography{references.bib}
\footnotesize

\end{document}